\documentclass[runningheads]{llncs}
\usepackage{amsmath}
\usepackage{marginnote}
\usepackage{graphicx}
\usepackage{url}
\usepackage{enumitem}
\usepackage{ulem}
\usepackage{soul,xcolor}
\usepackage{multirow}
\usepackage{multicol} 
\usepackage{threeparttable}
\usepackage{longtable}
\usepackage{csquotes}
\usepackage{lipsum}
\usepackage{amssymb}

\newcommand\blfootnote[1]{%
  \begingroup
  \renewcommand\thefootnote{}\footnote{#1}%
  \addtocounter{footnote}{-1}%
  \endgroup
}
\newcommand{\papertitle}{Neuro-Symbolic RDF and Description Logic Reasoners: The State-Of-The-Art and Challenges}
\begin{document}
\title{\papertitle}
%
%
\author{Gunjan Singh\inst{1}\and Sumit Bhatia\inst{2} \and Raghava Mutharaju\inst{1}}
\authorrunning{Singh et al.}
%
\institute{Knowledgeable Computing and Reasoning Lab, IIIT-Delhi, India \\
\email{\{gunjans, raghava.mutharaju\}@iiitd.ac.in}\\
\and
Media and Data Science Research Lab, Adobe Inc., Delhi, India\\
\email{sumit.bhatia@adobe.com}}
\maketitle              
\begin{abstract}
    Ontologies are used in various domains, with RDF and OWL being prominent standards for ontology development. RDF is favored for its simplicity and flexibility, while OWL enables detailed domain knowledge representation. However, as ontologies grow larger and more expressive, reasoning complexity increases, and traditional reasoners struggle to perform efficiently. Despite optimization efforts, scalability remains an issue. Additionally, advancements in automated knowledge base construction have led to the creation of large and expressive ontologies that are often noisy and inconsistent, posing further challenges for conventional reasoners. To address these challenges, researchers have explored neuro-symbolic approaches that combine neural networks' learning capabilities with symbolic systems' reasoning abilities. In this chapter, we provide an overview of the existing literature in the field of neuro-symbolic deductive reasoning supported by RDF(S), EL, ALC, and OWL 2 RL, discussing the techniques employed, the tasks they address, and other relevant efforts in this area.
\end{abstract}

\section{Introduction}

Ontologies are used to represent knowledge in several domains, such as
healthcare, geoscience, IoT, and e-commerce. In recent years, there has been an
uptake in expressing ontologies using the Resource Description Framework (RDF)\footnote{https://www.w3.org/TR/rdf-primer/} \cite{RDF} and
the Web Ontology Language (OWL)\footnote{https://www.w3.org/TR/owl2-primer/}~\cite{OWL2profile}, the prominent W3C recommended standards
for ontology development. The strength of these knowledge representation languages is automated reasoning, which can be used for various purposes, such as conceptual query answering, data retrieval, and data integration~\cite{Ontologies_and_the_Semantic_Web}.\blfootnote{} RDF, while having limited expressivity, is popular because of its simplicity and flexibility to model data in diverse domains~\cite{Heist2020KnowledgeGO}. \blfootnote{This paper is a part of the book titled \enquote{Compendium of Neuro-Symbolic Artificial Intelligence}  which can be found at the following link: \url{https://www.iospress.com/catalog/books/compendium-of-neurosymbolic-artificial-intelligence}}On the other hand, OWL, which is based on Description Logics (DLs)\footnote{Since there is a one-to-one correspondence between OWL and DLs, we use those two terms interchangeably here.}~\cite{dl_primer}, provides a range of language constructors (Table \ref{tab:syntax}) for capturing the domain knowledge in detail. 
 
Since RDF has limited expressivity, most of the commonly used reasoners\footnote{\url{http://owl.cs.manchester.ac.uk/tools/list-of-reasoners/}} perform quite well on these ontologies. However, as the ontologies grow larger or become more expressive, for example, the ones in OWL 2 DL~\cite{OWL2profile}, whose formal underpinning is the highly expressive description logic, $\mathcal{SROIQ}$~\cite{dl_sroiq}, the reasoning complexity also increases. The current reasoners struggle to perform efficiently on such ontologies. Modern reasoning systems typically employ a wide range of optimizations to improve performance. Based on the intended application, these optimizations may fall into different categories (pre-processing, model construction, and reasoning task-dependent optimizations)~\cite{dlbook}. However, despite all the research on optimizing the performance of traditional reasoning systems, they still suffer from poor scalability on large and
expressive ontologies~\cite{owl2bench}. Furthermore, with the advancements in automated knowledge base construction\footnote{\url{https://www.akbc.ws/}}, building large and expressive ontologies has become relatively easy. However, such ontologies are often noisy and inconsistent, and the conventional ontology reasoners cannot deal with such ontologies~\cite{neurosymbolic_AI}. 

To come up with a solution to deal with large, expressive, noisy, and inconsistent ontologies, researchers have explored neuro-symbolic approaches~\cite{neuro_symbolic} that combine the robust learning capabilities of the neural networks and the precise reasoning abilities of the symbolic systems. 
This chapter presents a survey of existing neuro-symbolic deductive reasoning techniques for RDF and description logics and discusses their benefits, limitations, and scope. Before we do so, let us point out that we exclude works that use machine learning techniques for knowledge graph completion (KGC)\footnote{\url{https://paperswithcode.com/task/knowledge-graph-completion}}, techniques for different ontology completion tasks, such as ontology learning for assisting ontology engineering, enrichment~\cite{ontology_population,Completing_DL_KB} and for introducing approximate class definitions that are not part of the standard ontology languages~\cite{concept_learning} and also predicting assertions in an ontology~\cite{Distance-based,PredictionofClassandPropertyAssertions}. 
Specifically, we discuss only those works that focus on deductive reasoning over ontologies. Such techniques infer only logically derivable facts from explicitly stated domain knowledge. The conventional deductive ontology reasoners are based on mathematical logic-based reasoning algorithms such as tableau calculi and inference rule-based methods (Section~\ref{conventional}).


This chapter is organized as follows. Section~\ref{preliminaries} gives a brief overview of ontology languages, followed by different traditional and neuro-symbolic reasoning techniques. Although there are several different description logics ($\mathcal{ALC}, \mathcal{SRIF}$, $\mathcal{SROIQ}$, etc.)~\cite{dl_sroiq}, and several OWL 2 profiles (EL, QL, RL, and DL)~\cite{OWL2profile}, the body of work in the neuro-symbolic reasoning category is limited. Hence, we only briefly discuss the ontology languages that are supported by neuro-symbolic reasoning techniques such as RDF(S), $\mathcal{EL}$, $\mathcal{ALC}$, and OWL 2 RL. 
Section~\ref{sota} discusses the existing literature in the neuro-symbolic deductive reasoning space. In section~\ref{summary}, we give a brief overview of the techniques discussed in the chapter in terms of the techniques used, tasks handled by each work, and so on. We also discuss a few complementary efforts in Section~\ref{other} that push the research in the ontology reasoning domain, such as reasoner evaluation challenges and ontology benchmarks. We conclude in Section~\ref{conclusion}.

\section{Preliminaries}
\label{preliminaries}
\subsection{Ontology Languages}
With the increasing interest in the Semantic Web~\cite{Sweb_foundation,semantic_web} in the early 2000s, ontologies~\cite{ontology} were used in various application domains such as healthcare, geoscience, IoT, and e-commerce to describe knowledge about entities in that particular domain. The basic building blocks of an ontology are entities: \texttt{Concepts (C)}, \texttt{Relations (R)}, and \texttt{Individuals (I)}. \texttt{Concepts} denote sets of individuals, \texttt{Relations} denote binary relations between individuals, and \texttt{individual} names denote a single individual in the domain. For example, an ontology representing the \texttt{University} domain might use concepts such as \texttt{Student} and \texttt{Faculty} to denote the set of all students and faculties in the university, relations such as \texttt{teaches} and \texttt{hasClassmate} to denote the binary relationships between faculty, course, and students, and individual names such as \texttt{alex} and \texttt{mary} to denote the individuals alex and mary. Note that, depending on the language, \texttt{Concept} may sometimes be referred to as \texttt{Class}, \texttt{Relation} as \texttt{Properties} or \texttt{Roles}, and \texttt{Individuals} as \texttt{Constants}. But, to keep the terminology consistent, we will stick to the terms \texttt{Concept}, \texttt{Relation}, and \texttt{Individual} throughout the survey (irrespective of the language addressed).
\newline\newline
\noindent
\textbf{Definition 1 (Ontology).} An Ontology \texttt{O}, is a set of statements, called axioms,  that is used to describe knowledge about a particular domain. These axioms can be categorized as a triplet = (\texttt{TBox}, \texttt{ABox}, \texttt{RBox}). Let $N_C$, $N_R$, and $N_I$
be countable, and pairwise disjoint sets of concept names, relation names, and individual names, respectively. \texttt{TBox} is the set of terminological axioms describing the relationships between named concept expressions/concepts, \texttt{C} $\in$ $N_C$. Every terminological axiom is in the form of \texttt{A} $\sqsubseteq$ \texttt{B} (concept inclusion) or \texttt{A} $\equiv$ \texttt{B} (concept equivalence), such that \texttt{A}, \texttt{B} $\in$ $N_C$.  \texttt{ABox} is the set of assertions describing  relationships among individuals \texttt{a}, \texttt{b}  $\in$ $N_I$ via relation \texttt{R} (relation assertion) $\in$ $N_R$  as well as instantiation relationships (concept assertion) between
elements of $N_I$ and $N_C$.  \texttt{RBox} is the set of relational axioms describing different properties of relations. 

RDF(S)~\cite{RDF} and OWL 2 (the latest version of OWL)~\cite{OWL2profile} are the two prominent W3C recommended standards for ontology development that provide several language constructors to model the information in a given domain. As mentioned earlier, OWL and DLs have a one-to-one correspondence. OWL 2 is based on DL $\mathcal{SROIQ}$. Table~\ref{tab:syntax} presents different subsets of constructs offered. Below are a few examples to distinguish between \texttt{TBox}, \texttt{ABox}, and \texttt{RBox} of an ontology.
\begin{itemize}
       \item \textbf{TBox}
      \begin{enumerate}
        \item[*] Class(\texttt{UGStudent})
        
        \textit{UG student is a concept}
        \item[*] \texttt{UGStudent} $\equiv$ \texttt{Student} $\sqcap$ $\exists$\texttt{enrollFor.Course}

        \textit{UG student is a student who enrolls in a UG Program}
        \item[*] \texttt{UGStudent} $\sqsubseteq$ \texttt{Student}

        \textit{UG student is a subclass of Student (rdfs:subClassOf)}
    \end{enumerate}
   \item \textbf{ABox}
    \begin{enumerate}
        \item [*]  \texttt{UGStudent}(\texttt{alex})

        \textit{alex is a UG Student (Concept Assertion)}
        \item [*] \texttt{teaches}(\texttt{mary}, \texttt{alex})
        
        \textit{mary teaches alex (Relation Assertion)}
    \end{enumerate}
    \item \textbf{RBox}
    \begin{enumerate}        
        \item[*] Symmetric(\texttt{hasClassmate})

        \textit{hasClassmate is a symmetric object property}
        \item[*]  Irreflexive(\texttt{teaches})

        \textit{teaches is an irreflexive object property}
    \end{enumerate}
\end{itemize} 

\begin{table}
\centering
\begin{tabular}{|cp{0.6cm}p{0.2cm}p{0.6cm}|p{2.0cm}|p{4.5cm}|p{2.5cm}|}
\hline
\multicolumn{4}{|c|}{\textbf{DL}} &
  \textbf{Syntax} &
  \textbf{Semantics} & 
  \textbf{Name} \\ \hline
\multicolumn{1}{|c|}{\multirow{22}{*}{\textbf{\rotatebox[origin=c]{90}{$\mathcal{SROIQ}$}}}} &
  \multicolumn{1}{c|}{\multirow{11}{*}{\textbf{$\mathcal{S}$}}} &
  \multicolumn{1}{c|}{\multirow{10}{*}{\textbf{$\mathcal{ALC}$}}} &
  \multirow{6}{*}{\textbf{$\mathcal{EL}$}} &
  C, \newline R, \newline a &
  $C^\mathcal{I} \subseteq \Delta^\mathcal{I}$, \newline $R^\mathcal{I} \subseteq \Delta^\mathcal{I}$ $\times$ $\Delta^\mathcal{I}$, \newline a $\in$ $\Delta^\mathcal{I}$ &
  Atomic (concepts, relation, individuals) \\ \cline{5-7} 
\multicolumn{1}{|c|}{} &
  \multicolumn{1}{c|}{} &
  \multicolumn{1}{c|}{} &
 &
  $\top$ &  $\Delta^\mathcal{I}$
   &
  Top \\ \cline{5-7} 
\multicolumn{1}{|c|}{} &
  \multicolumn{1}{c|}{} &
  \multicolumn{1}{c|}{} &
   &
  C(a), \newline R(a,b) &
  $a^\mathcal{I}$ $\in$ $C^\mathcal{I}$, \newline $\bigl \langle$$a^\mathcal{I}$, $b^\mathcal{I}$$\bigr \rangle$ $\in$ $R^\mathcal{I}$
   &
  Concept Assertion, Relation Assertion \\ \cline{5-7} 
\multicolumn{1}{|c|}{} &
  \multicolumn{1}{c|}{} &
  \multicolumn{1}{c|}{} &
   &
$ C \sqsubseteq D $, \newline$C \equiv D $ & 
   $C^\mathcal{I} \subseteq D^\mathcal{I}$ , \newline $C^\mathcal{I} = D^\mathcal{I}$ 
   &
  Concept Inclusion, Concept Equivalences \\ \cline{5-7} 
\multicolumn{1}{|c|}{} &
  \multicolumn{1}{c|}{} &
  \multicolumn{1}{c|}{} &
   &
$  C \sqcap D $ &  $C^\mathcal{I} \cap D^\mathcal{I}$ 
   &
  Conjuction \\ \cline{5-7} 
\multicolumn{1}{|c|}{} &
  \multicolumn{1}{c|}{} &
  \multicolumn{1}{c|}{} &
   &
 $\exists R.C$ & $\{$a $\in \Delta^\mathcal{I} |$ there is a $\bigl \langle$a, b$\bigr \rangle$ $  \in R^\mathcal{I} $ with  $ b \in C^\mathcal{I}) \}$
  
   &
  Existential Restriction \\ \cline{4-7} 
\multicolumn{1}{|c|}{} &
  \multicolumn{1}{c|}{} &
  \multicolumn{1}{c|}{} &
  \textbf{} &
 $ \bot$ & $\phi$
   &
  Bottom \\ \cline{4-7} 
\multicolumn{1}{|c|}{} &
  \multicolumn{1}{c|}{} &
  \multicolumn{1}{c|}{} &
  \textbf{} &
  $\neg$ C & $\Delta^\mathcal{I} \backslash C^\mathcal{I}$
   &
  Negation \\ \cline{4-7} 
\multicolumn{1}{|c|}{} &
  \multicolumn{1}{c|}{} &
  \multicolumn{1}{c|}{} &
  \textbf{} &   
  $C \sqcup D $ & $C^\mathcal{I} \cup D^\mathcal{I}$
   &
  Disjunction \\ \cline{4-7} 
\multicolumn{1}{|c|}{} &
  \multicolumn{1}{c|}{} &
  \multicolumn{1}{c|}{} &
  \textbf{} &
  $\forall R. C $ & $\{$a $\in \Delta^\mathcal{I} |$ if $\bigl \langle$a, b$\bigr \rangle \in R^\mathcal{I} $, then  $ b \in C^\mathcal{I}) \}$
   &
  Universal Restriction \\ \cline{3-7} 
\multicolumn{1}{|c|}{} &
  \multicolumn{1}{c|}{} &
  \multicolumn{1}{c|}{\multirow{7}{*}{\textbf{$\mathcal{R}$}}} &
  \textbf{} &
  Trans(R) & if $\bigl \langle$a, b$\bigr \rangle$ $  \in R^\mathcal{I}$ for all a,b,c $\in \Delta^\mathcal{I}$ for which $\bigl \langle$a, b$\bigr \rangle$ $ \in R^\mathcal{I}$ and   $\bigl \langle$b, c$\bigr \rangle$ $  \in R^\mathcal{I}$
   & 
  Transitivity \\ \cline{2-2} \cline{4-7} 
\multicolumn{1}{|c|}{} &
  \multicolumn{1}{c|}{} &
  \multicolumn{1}{c|}{} &
  \textbf{$\mathcal{H}$} &
$R \sqsubseteq S$,\newline $R\equiv S $&
  $R^\mathcal{I} \subseteq S^\mathcal{I}$ , \newline $R^\mathcal{I} = S^\mathcal{I}$ 
   &
  Relation Inclusions / Equivalences \\ \cline{2-2} \cline{4-7} 
\multicolumn{1}{|c|}{} &
  \multicolumn{1}{c|}{} &
  \multicolumn{1}{c|}{} &
   &
  R1 o R2 $\sqsubseteq S $& $R^\mathcal{I}$ o $R2  \subseteq S^\mathcal{I}$
   & 
  Relation Composition \\ \cline{2-2} \cline{4-7} 
\multicolumn{1}{|c|}{} &
  \multicolumn{1}{c|}{} &
  \multicolumn{1}{c|}{} &
   &
  U & $\Delta^\mathcal{I}$ $\times$ $\Delta^\mathcal{I}$
   &
  Universal Relation \\ \cline{2-2} \cline{4-7} 
\multicolumn{1}{|c|}{} &
  \multicolumn{1}{c|}{} &
  \multicolumn{1}{c|}{} &
   & 
  Refl(R), \newline Symm(R),... & if $\bigl \langle$a, a$\bigr \rangle$ $  \in R^\mathcal{I}$ for all a $\in \Delta^\mathcal{I}$ , \newline if $\bigl \langle$a, b$\bigr \rangle$ $  \in R^\mathcal{I}$ for all $\bigl \langle$b, a$\bigr \rangle$ $ \in R^\mathcal{I}$
   &
  Reflexivity, Symmetry, ... \\ \cline{2-2} \cline{4-7} 
\multicolumn{1}{|c|}{} &
  \multicolumn{1}{c|}{} &
  \multicolumn{1}{c|}{} &
   &
  Disj(R, S) & $R^\mathcal{I} \cap S^\mathcal{I} = \phi$
   &
  Disjoint Roles \\ \cline{2-2} \cline{4-7} 
\multicolumn{1}{|c|}{} &
  \multicolumn{1}{c|}{} &
  \multicolumn{1}{c|}{} &
   &
  $\exists $R. Self & $\{$a $\in \Delta^\mathcal{I}$  $ \mid \bigl \langle$a, b$\bigr \rangle$ $  \in R^\mathcal{I} $\}
   &
  Self Restrictions \\ \cline{2-7} 
\multicolumn{1}{|c|}{} &
  \multicolumn{3}{c|}{\textbf{$\mathcal{O}$}} &
  \{a\} & \{$a^\mathcal{I}$\}
   &
  Nominals \\ \cline{2-7} 
\multicolumn{1}{|c|}{} &
  \multicolumn{3}{c|}{\textbf{$\mathcal{I}$}} &
 $ R^-$ & \{$\bigl \langle$b, a$\bigr \rangle$ $\mid$ $\bigl \langle$a, b$\bigr \rangle \in R^\mathcal{I}$\}
   &
  Inverse Relation \\ \cline{2-7} 
\multicolumn{1}{|c|}{} &
  \multicolumn{1}{c|}{\multirow{3}{*}{\textbf{$\mathcal{Q}$}}} &
  \multicolumn{1}{c|}{\multirow{2}{*}{\textbf{$\mathcal{N}$}}} &
  \textbf{$\mathcal{F}$} &
  Func(R), \newline$ \leq$ 1. R & if a=b for all a, b, c $\in \Delta^\mathcal{I}$ for which $\bigl \langle$a, b$\bigr \rangle$ $\in R^\mathcal{I}$ and $\bigl \langle$a, c$\bigr \rangle$ $\in$ $R^\mathcal{I}  $
   &
  Functional Relation/ Functional Cardinality Restrictions \\ \cline{4-7} 

\multicolumn{1}{|c|}{} &
  \multicolumn{1}{c|}{} &
  \multicolumn{1}{c|}{} &
   &
 $ \geq $n R. C, \newline$ \leq $ n R. C & \{a $\in \Delta^\mathcal{I} \mid $  \{b $\in \Delta^\mathcal{I}$  $ \mid \bigl \langle$a, b$\bigr \rangle$  $\in$ $R^\mathcal{I}$ and b $\in C^\mathcal{I}\} \geq$ n \} \newline \{a $\in \Delta^\mathcal{I} \mid $  \{b $\in \Delta^\mathcal{I}$  $ \mid \bigl \langle$a, b$\bigr \rangle$  $\in$ $R^\mathcal{I}$ and b $\in C^\mathcal{I}\} \leq$ n \}
   &
  Qualified At least / \newline At most Cardinality Restrictions \\ \hline
\end{tabular}
\caption{Syntax and Semantics of the description logic $\mathcal{SROIQ}$}
\label{tab:syntax}
\end{table}

The Resource Description Framework (RDF), which includes RDF Schema (RDFS), is an established and widely used W3C standard for expressing knowledge graphs. 
An RDF graph, \texttt{G},  is a directed, labeled graph that consists of a set of triples of the form {(\texttt{s}, \texttt{p}, \texttt{o})} where the subject \texttt{s} and object \texttt{o} form the graph nodes and the predicate \texttt{p} forms the graph edges, where subject \texttt{s} $\in$ $N_C$ or $N_I$, and object \texttt{o} $\in$  $N_C$, $N_I$ or \texttt{Literals}, and the predicate \texttt{p} $\in$ $N_R$ denote the binary relation between the nodes \texttt{s} and \texttt{o}. 

RDF Schema (RDFS) is a semantic extension of basic RDF that provides several constructs such as \texttt{rdfs:subClassOf}, \texttt{rdfs:subPropertyOf}, \texttt{rdfs:Class}, \texttt{rdfs:Resource}, \texttt{rdfs:Literal}, and \texttt{rdfs:Datatype} to express simple taxonomies and hierarchies among entities. RDF(S) can be used to create lightweight ontologies, and the information in an RDFS ontology can also be represented as a directed, labeled graph. Note that we use RDF(S) to refer to both RDF and RDFS.   


OWL, built on top of RDFS, offers more language constructs. The basic building blocks of OWL are concepts (C), relations (R), and individuals (I), which can be put into relationships (called axioms) with each other using different modeling constructs, as shown in Figure \ref{tab:syntax}. The sets C, R, and I must be mutually disjoint. All or a set of these constructs could be utilized depending on the intended application and desired reasoning performance. 

OWL 2~\cite{owl2}, the latest version of OWL, is based on DL $\mathcal{SROIQ}$. Since DLs are a family of first-order logics (FOL)~\cite{fol}, formal semantics are defined for this language. An interpretation $\mathcal{I}$ = ($\Delta^\mathcal{I}, .^\mathcal{I}$), where $\Delta^\mathcal{I}$, a non-empty set, called the domain of $\mathcal{I}$ (representing all the things existing in the world that $\mathcal{I}$ represent), and $.^\mathcal{I}$ is the interpretation function that maps every \texttt{individual}, \texttt{concept}, and \texttt{relation} in $N_I$, $N_C$,  and $N_R$,  to an element, subset and binary relation in $\Delta^\mathcal{I}$ with function $.^\mathcal{I}$. For an interpretation $I$, the function $.^\mathcal{I}$ is defined in (Table~\ref{tab:syntax}).

OWL 2  has several sublanguages or profiles (EL, QL, RL, DL, and Full)~\cite{OWL2profile}, which are overlapping subsets of DL language constructs. In simpler terms, an OWL ontology could be referred to in terms of one of these profiles (for instance, OWL 2 DL ontology) or a (subset) combination of different DL constructs (such as $\mathcal{SHIQ}$, $\mathcal{ALCH}$, or $\mathcal{SHOIN}$). 

Different DL constructs (Table~\ref{tab:syntax})  have different computational properties. Some are easier
to reason over than others. Since the reasoning performance decreases with an increase in
expressivity, the primary purpose of the different DLs (formed by various combinations of constructs) is that they enable a balance between 
ontology expressivity and reasoning complexity depending on the intended application.  
For example, a lightweight DL, $\mathcal{EL}$, provides only a few concept constructors, which can be reasoned over in polynomial time w.r.t. the size of the ontology, even in the worst case. Two extensions of $\mathcal{EL}$, $\mathcal{EL}^{+}$, and $\mathcal{EL}^{++}$ support a few more concept constructors to make the ontology more expressive. $\mathcal{EL}^{+}$ additionally supports relation inclusion and composition. On top of that, $\mathcal{EL}^{++}$ further includes concept disjointness, transitive and reflexive relations, along with concept and role assertions.  
So, for applications where reasoning time is a significant concern, such DLs enable efficient implementations for even large ontologies. More expressive DLs (often termed semi-expressive DLs), such as $\mathcal{ALC}$, allow for including  conjunction, disjunction, negation, existential and universal quantification in the domain description. 
Finally, highly expressive DL $\mathcal{SROIQ}$~\cite{dl_sroiq} provides an extensive range of language constructs to represent an application domain. Reasoning over such expressive DLs is typically less efficient compared to simpler ones. However, the additional expressiveness can be helpful (and, in some cases, required) in practice to accurately represent the application domain's constraints. 

Each OWL 2 profile~\cite{owl2} consists of different (overlapping) subsets of DL constructs and hence vary in terms of their expressivity and reasoning complexity. 
\begin{itemize}
    \item \textbf{OWL 2 EL} (Existential Logic). Based on the DL $\mathcal{EL++}$ and designed to be lightweight and allow polynomial time reasoning.
    \item \textbf{OWL 2 QL} (Query Logic). Based on DL-Lite family of DLs and is designed to give easier access to a large amount of instance data.
    \item \textbf{OWL 2 RL} (Rules Logic). Based on DLP and pD* and designed to interact with rule-based reasoners.
    \item \textbf{OWL 2 DL} (Description Logics). Highly expressive, based on $\mathcal{SROIQ}$ and has the reasoning complexity of N2EXPTIME. 
    \item \textbf{OWL 2 Full.} Undecidable.
\end{itemize} 

Each of these profiles has some restrictions in terms of syntax. To illustrate the syntactic limits imposed on the usage of different constructs in each OWL 2 profile,  we use existential restriction for the statement, \textit{A faculty is an employee who teaches some course}. The syntax in each OWL 2 profile varies, as shown below.

\begin{itemize}
  \item In OWL 2 EL, \texttt{Faculty} $\equiv$ \texttt{Employee} $\sqcap$ $\exists$\texttt{teaches.Course}
  \item In OWL 2 QL, \texttt{Faculty} $\sqsubseteq$ \texttt{Employee} $\sqcap$ $\exists$\texttt{teaches.Course} (existential restrictions are not allowed in the subclass expression)
  \item In OWL 2 RL, \texttt{Employee}  $\sqcap$ $\exists$\texttt{teaches.Course} $\sqsubseteq$ \texttt{Faculty} (existential restrictions are not allowed in the superclass expression)   
  \item In OWL 2 DL, \texttt{Faculty} $\equiv$ \texttt{Employee} $\sqcap$ =1\texttt{teaches.Course} (since qualified exact cardinalities are supported, we could make the axiom more expressive by writing it using exact cardinality)
\end{itemize} 

\subsection{Reasoning}
The conventional ontology reasoning methods employ mathematical algorithms, such as tableau and inference rule-based methods, to perform different reasoning tasks. These algorithms, in contrast to learning-based approaches, are capable of deriving every possible conclusion (completeness) with absolute certainty (soundness) in a finite amount of time (termination). However, the conventional methods, despite a range of optimizations, need to scale better for large and expressive ontologies. In addition, these algorithms assume the integrity of the data and can not deal with noisy and inconsistent ontologies. Hence, neuro-symbolic techniques come into the picture because they are often highly scalable and have the added advantage of dealing with noise and providing robust learning abilities. 

Before moving on to the existing neuro-symbolic literature, it is crucial to understand the terminologies involved and what exactly (such as conventional reasoning methods or tasks) the neuro-symbolic reasoners try to emulate. This section briefly discusses the most relevant conventional reasoning methods for RDF(S) and DLs, followed by a general idea of neuro-symbolic reasoning in the context of this survey.


\subsubsection{Conventional Reasoning Methods}
\label{conventional}
The existing reasoning systems support a range of reasoning tasks. Depending on the intended application, these systems could be query-based, which tells whether a given logical expression is a logical consequence of the provided ontology or generates all the deductive inferences in a single run. For the latter case, the tasks that are typically considered standard for ontology reasoning and are supported by most reasoning systems are as follows. 
\begin{itemize}
    \item \textbf{Consistency Checking}. Determines whether the given set of axioms is consistent, i.e., no entity violates or contradicts the given ontology definition.
    \item \textbf{Satisfiability}. Determines whether a description of the concept is not contradictory, i.e., checks whether at least one individual satisfies the given concept description.
    \item \textbf{Classification}. Determines the subsumption hierarchies for all the concepts and relations.
    \item \textbf{Realization}.  Determines all the concepts to which the individual belongs (often called Concept Membership), especially the most specific ones.
    \item \textbf{Entailment}. Determines all the logical consequences (including classification and realization) that follow from the given ontology, provided the ontology is consistent. 
\end{itemize}
From the point of view of conventional reasoners, the following aspects are typically investigated to measure the effectiveness of the implemented reasoning algorithms.
\begin{itemize}
    \item \textbf{Soundness}. A procedure is called sound if it derives only those consequences that logically follow from the represented axioms.
    \item \textbf{Completeness}.  A procedure is called complete if the procedure can derive all the logical consequences that follow from the given ontology.
    \item \textbf{Termination}. A procedure should  compute sound and complete answers in a finite amount of time. 
    \item \textbf{Scalability}. Indicates how the reasoner performs for varying sizes of ontology (in terms of axiom count and type of language constructs used). The performance is typically measured in terms of reasoning time taken and memory consumed.
\end{itemize} 
Both soundness and completeness are strongly desired features, but in some cases, completeness can be compromised to obtain faster results.  

\paragraph{\textbf{Tableau-Based Methods.}} For DLs, all the reasoning tasks are reducible to consistency checking. Hence it is sufficient to have a procedure that only determines whether or not ontologies
are consistent. 
Let C be a concept, and I be an interpretation. The aim of the tableau algorithm~\cite{dl_sroiq} is to generate a finite interpretation I such that $C^I \neq  \emptyset$. 
Without loss of generality, we assume all concept descriptions are in the negation normal form (NNF)~\cite{Sweb_foundation}. 
\begin{figure}[ht]
\centering 
     \includegraphics[width=0.80\textwidth]{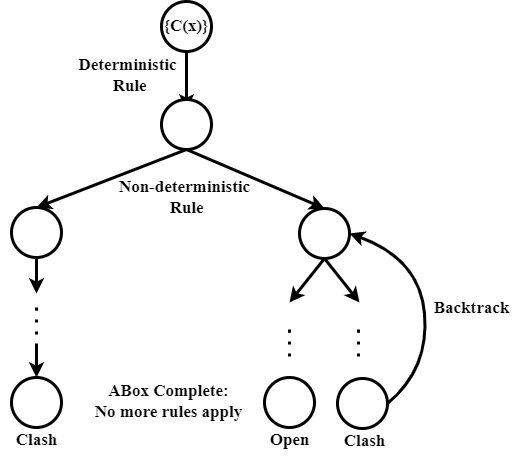}

       \caption{Tableau Search Tree}
         \label{fig:tab}
      
 \end{figure}
Tableau expansion starts with an ontology, A:= {C(x)}, and the tableau expansion rules are applied sequentially until all constraints are satisfied, or an obvious contradiction is detected. 
This leads to a form of a search tree (Figure ~\ref{fig:tab}), with the root node {C(x)}. Here, the edges represent the applied tableau rules. 
We call our ontology A consistent if and only if one of the complete ABoxes is open, i.e., does not contain an obvious contradiction of the form $\{A(x), \neg A(x)\}$. These rules may be deterministic (such as conjunction) or non-deterministic (such as disjunction). The second type of rule might lead to the wrong path and require backtracking. Thus, with the increase in the size or expressivity of the ontologies, the tree becomes broader and deeper, leading to a huge search space. Hence, the reasoning complexity increases exponentially. 

\paragraph{\textbf{Inference Rule-Based Methods.}}
Inference rule-based reasoners are widely used and well-known for bottom-up reasoning techniques that derive all the inferences by applying inference rules to the set of axioms in the input ontology iteratively (repeatedly). There are thirteen rules (called entailment rules) for RDFS (Table ~\ref{tab:rdfsrules}) that can be used to entail new facts in an RDFS ontology.  Since RDFS has very low expressivity and axioms are in the form of triples, reasoning algorithms are very straightforward. However, for $\mathcal{EL}^{+}$, the axioms are more complex than RDFS. So, usually, some preprocessing has to be done before applying the rules, such as all the axioms have to be in the standard normal form: $  C \sqsubseteq D,  C_1 \sqcap C_2 \sqsubseteq D,    C \sqsubseteq \exists R. D,  \exists R.  \sqsubseteq  D,  R_1 \sqcap R,  R_1$ o $R_2 \sqcap R $. The standard reasoning task over $\mathcal{EL+}$ is called classification. The inference rules (called completion rules) (Table~\ref{tab:elrules}) can be applied to $\mathcal{EL}^{+}$ axioms in a given $\mathcal{EL}^{+}$ ontology to derive all the possible consequences. These approaches have also been extended to DLs with nondeterministic language features, e.g., $\mathcal{ALCH}$ and $\mathcal{ALCI}$ (refer to~\cite{consequence_DL} for more details). 
\begin{table}
\begin{center}
\begin{tabular}{|p{1.2cm}|p{5.2cm}|p{1.2cm}|p{4.2cm}|}
\hline
\textbf{Rule}   & \textbf{Premise}                                                                                       & \textbf{Pattern Type}                                       & \textbf{Conclusion}                         \\ \hline
RDFS1  & any IRI aaa in D                                                                                  & -                                                   & aaa rdf:type rdfs:Datatype.         \\ \hline
RDFS2  & \begin{tabular}[c]{@{}l@{}}aaa rdfs:domain xxx.\\ yyy aaa zzz.\end{tabular}                       & \begin{tabular}[c]{@{}l@{}}TBox\\ ABox\end{tabular} & yyy rdf:type xxx.                   \\\hline
RDFS3  & \begin{tabular}[c]{@{}l@{}}aaa rdfs:range xxx.\\ yyy aaa zzz.\end{tabular}                        & \begin{tabular}[c|]{@{}l@{}}TBox\\ ABox\end{tabular} & zzz rdf:type xxx.                   \\\hline
RDFS4a & xxx aaa yyy.                                                                                      & -                                                   & xxx rdf:type rdfs:Resource.         \\\hline
RDFS4b & xxx aaa yyy.                                                                                      & -                                                   & yyy rdf:type rdfs:Resource.         \\\hline
RDFS5  & \begin{tabular}[c]{@{}l@{}}xxx rdfs:subPropertyOf yyy.\\ yyy rdfs:subPropertyOf zzz.\end{tabular} & \begin{tabular}[c]{@{}l@{}}TBox\\ TBox\end{tabular} & xxx rdfs:subPropertyOf zzz.         \\\hline
RDFS6  & xxx rdf:type rdf:Property                                                                         & TBox                                                & xxx rdfs:subPropertyOf xxx.         \\\hline
RDFS7  & \begin{tabular}[c]{@{}l@{}}aaa rdfs:subPropertyOf bbb.\\ xxx aaa yyy.\end{tabular}                & \begin{tabular}[c]{@{}l@{}}TBox\\ ABox\end{tabular} & xxx bbb yyy.                        \\\hline
RDFS8  & xxx rdf:type rdfs:Class                                                                           & TBox                                                & xxx rdfs:subClassOf rdfs:Resource   \\\hline
RDFS9  & \begin{tabular}[c]{@{}l@{}}xxx rdfs:subClassOf yyy.\\ zzz rdf:type xxx.\end{tabular}              & \begin{tabular}[c]{@{}l@{}}TBox\\ ABox\end{tabular} & zzz rdf:type yyy.                   \\\hline
RDFS10 & xxx rdf:type rdfs:Class.                                                                          & TBox                                                & xxx rdfs:subClassOf xxx.            \\\hline
RDFS11 & \begin{tabular}[c]{@{}l@{}}xxx rdfs:subClassOf yyy.\\ yyy rdfs:subClassOf zzz.\end{tabular}       & \begin{tabular}[c]{@{}l@{}}TBox\\ TBox\end{tabular} & xxx rdfs:subClassOf zzz.            \\\hline
RDFS12 & xxx rdf:type  rdfs:ContainerMembershipProperty.                               & TBox                                                & xxx rdfs:subPropertyOf rdfs:member. \\\hline
RDFS13 & xxx rdf:type rdfs:Datatype.                                                                       & TBox                                                & xxx rdfs:subClassOf rdfs:Literal.  \\
\hline
\end{tabular}
\end{center}
\caption{RDF/RDFS Entailment Rules}
\label{tab:rdfsrules}
\end{table}

\begin{table}
\begin{center}
\begin{tabular}{|c|ccccc|}
\hline
1 
& A $\sqsubseteq$ C 
& C $\sqsubseteq$ D 
&  
& $\vDash$ 
& A $\sqsubseteq$ D 
\\ \hline
2
& A $\sqsubseteq$ $C_1$ 
&  A $\sqsubseteq$ $C_2$ 
& $C_1 \sqcap C_2$ $\sqsubseteq$ D 
& $\vDash$ 
&A$\sqsubseteq$ D
\\ \hline
3
& A $\sqsubseteq$ C 
&  C $\sqsubseteq \exists$R.D 
&  
& $\vDash$ 
&A$\sqsubseteq\exists$R.D 
\\ \hline
4& A $\sqsubseteq \exists$R.B & B $\sqsubseteq$ C & $\exists$R.C $\sqsubseteq$ D  & $\vDash$ &A $\sqsubseteq$ D \\ \hline
5& A $\sqsubseteq \exists$S.D & S $\sqsubseteq$ R &  & $\vDash$ &A $\sqsubseteq \exists$R.D \\ \hline
6&  A $\sqsubseteq\exists R_1.C $ &  C $\sqsubseteq\exists   R_2.D$ & $R_1$ o $R_2 \sqsubseteq$ R  & $\vDash$ &A $\sqsubseteq \exists$R.D \\ \hline
\end{tabular}
\end{center}
\caption{$\mathcal{EL+}$ Completion Rules}
\label{tab:elrules}
\end{table}

An important point to note is that for the conventional logical reasoning algorithms, the entities in an ontology need not necessarily have an actual (real-world) meaning. The reasoner processes the axioms based on the defined rules and has nothing to do with entity names. For example, in inference-based methods, all the inferences are derived by the application of rules. For the axioms, \texttt{Man} $\sqsubseteq$ \texttt{Mortal} and \texttt{Man}(\texttt{socrates}), imply \texttt{Mortal}(\texttt{socrates}). If we replace \texttt{Man}, \texttt{socrates}, and \texttt{Mortal} with random names \texttt{A}, \texttt{x}, and \texttt{B}, respectively, then, \texttt{A} $\sqsubseteq$ \texttt{B} and \texttt{A}(\texttt{x}) will imply \texttt{B}(\texttt{x}). Even for the case of tableau-based reasoners, a consistent renaming across the whole ontology would not change the outcome of the reasoning task at hand. 

\subsubsection{Neuro-symbolic Reasoning Methods}
\label{neuro}
Henry Kautz, in his AAAI 2020 Robert S. Engelmore Memorial Award Lecture, discussed five categories of neuro-symbolic AI systems as the ``Future of AI"\footnote{\url{https://www.youtube.com/watch?v=_cQITY0SPiw}, starting at minute 29}. However, the goal of neuro-symbolic approaches, in the context of this survey, is to devise a model based on deep neural networks that can perform logical reasoning by learning the functionality of basic ontology reasoners. This could be done by learning the embeddings of the entities involved in the ontology or by learning the reasoning steps followed by the conventional tableau or inference rule-based systems. In contrast to conventional reasoning approaches, these approaches aren't sound and complete. They work by finding a balance between approximating the precise reasoning abilities of the symbolic systems and utilizing the robust learning capabilities of learning-based techniques.

From the point of view of evaluating a neuro-symbolic reasoning procedure, the key aspects ~\cite{pointer_nw}, in addition to the previously discussed conventional reasoning evaluation measures, are as follows.
\begin{itemize}
    \item \textbf{Transferability}. Indicates whether the system can transfer the knowledge learned during training to previously unknown and unseen ontology domains. As in the case of conventional logic-based algorithms, the reasoning systems are independent of the meaning of the entities in the ontology. Although the knowledge is not transferable for conventional reasoners, the same set of rules and reasoning algorithms are applicable across domains. The goal is that the neuro-symbolic reasoner should be able to learn the abstraction irrespective of the textual meaning of the entities, and hence, the learned knowledge can be transferred to new ontologies and would not need retraining. 
    \item \textbf{Generative Capability}. Indicates whether the system can generate all inferences in one run (in a finite amount of time) or if it is query-based, i.e., predicts the answer to the specific query or checks whether or not a given logical expression is a logical consequence of the provided ontology.
    \item \textbf{Transparency}. The ability of the neuro-symbolic systems to provide a derivation of the inferred triples and provide an explanation for the predictions. Neural networks are generally black boxes and cannot be inspected. So, this aspect indicates whether the proposed method incorporates the transparency offered by the conventional reasoning methods.
    \item \textbf{Accuracy}. The performance of a neuro-symbolic reasoning system, in contrast to conventional methods that are both sound and complete, is normally measured in terms of the number of correct predictions made for each query or correct inferences drawn (such as precision, recall, or f-measure) from an input ontology.
\end{itemize}

\paragraph{\textbf{Logical Embeddings.}}

Recent years have witnessed the successful application of low-dimensional vector space representations of knowledge graphs~\cite{KG} for the knowledge graph completion tasks, such as predicting missing links and new facts ~\cite{survey_KGR1,survey_KGR2}. The primary goal is to capture and preserve the underlying properties of the given knowledge base in the vector form. However, it is not yet well understood to what extent ontological knowledge, e.g., given as a set of axioms, can be embedded in a principled way. The goal of logical embeddings for ontologies is two-fold. 
\begin{enumerate}
    \item  The learned embedding should disregard the textual meanings of all the entities in the ontology and retain the syntax (type and order of constructs), and capture the interconnection and interplay between the axioms.
    \item The learning should be transferable either by learning the embedding for each entity and using those for ontology completion or query answering across domains or by learning the functioning of each construct involved in the ontology. The latter case is preferred for emulating conventional logical reasoning. 
\end{enumerate}

\paragraph{\textbf{Intelligent Decision Making.}}

\begin{figure}[ht]
\centering 
     \includegraphics[width=1.00\textwidth]{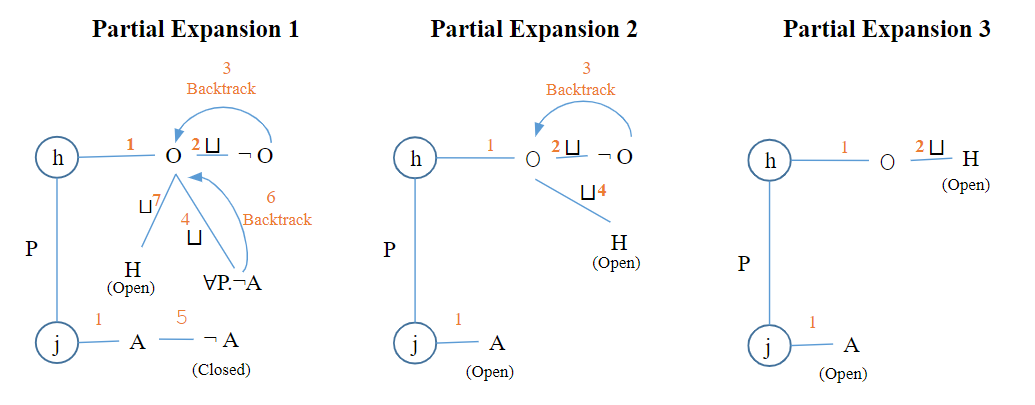}

       \caption{Partial tableau expansions for the ontology consisting of axioms, $\neg$ H $\sqcup \exists$ P.H, $\neg$ O $\sqcup$ (H $\sqcup \forall $P .$\neg$ A), O(h), P(h,j), A(j) }
         \label{fig:2}
      
 \end{figure}

Despite all the research for optimizing the performance of traditional reasoning systems, they still do not scale well. One of the reasons is that a tableau expansion still has a large degree of freedom in deciding which rule on which node should be applied next, as well as in which order it wants to evaluate the nondeterministic alternatives. The former choice does not make a huge difference because if there is a model, it will be found by any order of rule expansions. However, for the latter case, the choice is relevant because out of available choices, one choice may lead to the successful construction of a model, while another might lead to a clash and will have to be dealt with by a backtracking search. In the worst case, such a search must consider all non-deterministic choices and this can have a tremendous impact on the size of the search tree and, therefore, on the reasoning performance. Figure~\ref{fig:2} shows different partial expansions for a single set of axioms~\cite{Sweb_foundation}. Numbers indicate the order in which the tableau is expanded. Out of these three expansions, partial expansion 3 is the ideal choice for a reasoner because it took the least steps to reach an open branch. 
So, the ideal heuristic is one that will lead to the conclusion in the least number of steps. But, it is more important not to choose a bad heuristic because a bad decision might lead to a very broad and deep search tree, degrading the reasoning performance tremendously. 

\section{Neuro-Symbolic Reasoning: State-of-the-art}
\label{sota}

In this article, by examining the state-of-the-art, we plan to answer the following research questions.
\begin{enumerate}
    \item [Q1] \textit{Can neural networks learn to emulate logical reasoning?}
    
    An answer to this question is, in turn, dependent on the following questions.   
    \begin{enumerate}
    \item [Q1.1] \textit{What are the different neural network architectures and the corresponding algorithms to learn logical reasoning?}
    \item [Q1.2] \textit{How can ontologies and, in general, logical axioms and constructs be represented in a form suitable for different neural network architectures and algorithms?}
    \end{enumerate} 

    \item [Q2] \textit{How capable are the learned models in comparison to conventional RDF and description logic reasoners?}
    \begin{enumerate}
    \item [Q2.1] \textit{Can they handle noisy and inconsistent ontologies?}
    \end{enumerate}
    An ideal neuro-symbolic reasoner should support the most expressive logic, be transferable to different domains, be able to generate all and only correct inferences in a single run, and explain the generated consequences at a massive scale and with high performance. So the other important question is the following.
    \begin{enumerate}
    \item [Q2.2] \textit{To what extent can neuro-symbolic approaches support these features that were also discussed in Section~\ref{neuro}?} 
    
    In order to evaluate these features, it is important to have benchmarks that can be used to test and compare different neuro-symbolic reasoning systems.
    \end{enumerate}

\item [Q3] \textit{What are the metrics that can be used to evaluate the features of the neuro-symbolic reasoning systems?}
\end{enumerate}
\subsection{RDF}
\label{rdf}
Since RDF data can be represented as directed, labeled graphs, this information in machine learning scenarios has been utilized in two different forms -- general graph kernels~\cite{general_graph_kernel} and vector space embeddings~\cite{word2vec}. With graph-based representations, other than the benefit of improved readability, embeddings can capture the neighborhood information (context) of each entity. So the most intuitive form to use is representing ontology as a graph that can capture the syntax, but this is a complex task. Several graph models (such as bipartite, hypergraph, and metagraph)~\cite{bipartite,hypergraph,metagraph} have been explored for different purposes ranging from storing and querying RDF graphs to reducing space and time-complexity to solving the reification and provenance problem.~\cite{noise-tolerant} Unfortunately, they are not suitable for RDFS reasoning, and these graph forms cannot be provided as input to neural networks because the complex graph forms make it harder for the neural networks to learn from. 

Kernel methods refer to machine learning algorithms that learn by comparing pairs of data points using similarity (or dissimilarity) measures and work directly on graphs without having to do feature extraction to transform them into fixed-length, real-valued feature vectors. The distance between two data instances is computed by counting common substructures in the graphs of the instances, i.e., walks, paths, and trees~\cite{rdf2vec}. However, only a few approaches are general
enough to be applied to any given RDF graph, let alone OWL ontologies. In one of the earliest works in this category, L{\"o}sch et al. ~\cite{graph_kernel} introduced two general RDF graph kernels based on intersection graphs and intersection trees. Later, de Vries et al.~\cite{fast_graph_wfl_kernel} simplified the intersection tree path kernel and proposed a better and faster variant for RDF of the general Weisfeiler-Lehman kernel for graphs. In another work, de Vries et al.~\cite{fast_and_simple_graph_kernel} showed an improvement in the computational time of the kernel when applied to RDF by introducing an approximation of the state-of-the-art Weisfeiler-Lehman graph kernel algorithm. 

\begin{figure}[ht]
\centering 
     \includegraphics[width=1.00\textwidth]{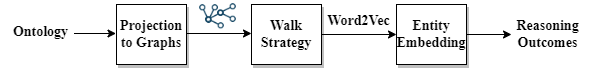}

       \caption{Ontology to vector representation pipeline}
         \label{fig:3}
      
 \end{figure}

In the vector space embedding, the initial works adapted neural language models, such as Word2Vec~\cite{word2vec}, for embedding entities present in the input ontology. Such approaches work on the assumption that closer words in the word sequences are more associated. As shown in Figure~\ref{fig:3}, the main step is to generate sequences that can be considered 
similar to sentences in natural language. These sequences could either be obtained directly from the axioms~\cite{opa2vec,onto2vec} or generated from the ontology graphs ~\cite{owl2vecstar}. Evidently, for the latter case, it is important to have a graphical representation of the ontologies. Once we have the graphs corresponding to the input ontology, several sequences are generated, and these sequences are used to train language models to generate the entity embeddings that capture the likelihood of a sequence of entities. After the training is completed, all the entities  are projected into a lower-dimensional feature space, and semantically similar entities are positioned close to each other. These word embeddings can be exploited for tasks related to query answering, ontology completion (such as link prediction and fact prediction), and ontology reasoning (such as predicting concept membership and subsumption). 

The initial work, RDF2Vec~\cite{rdf2vec} by Ristoski et al., leverages graph walks for transforming the RDF graph to token sequences, and then a natural language embedding algorithm such as Word2Vec~\cite{word2vec} is applied to the sequence document that generates embeddings based on token co-occurrences. Word2vec is a computationally-efficient two-layer neural network model
for learning word embeddings from raw text. For generating the sequences, two approaches were used; a graph walk based on the breadth-first algorithm and Weisfeiler-Lehman subtree RDF graph kernel, an efficient kernel for graph comparison~\cite{wf_lehman_kernel}.  The RDF2vec evaluations prove the feasibility and scalability of the graph walk-based methods over kernel methods. Since the embeddings are created using language models, the language(RDF) semantics are ignored. However, RDF2Vec could be used to study the distributional properties of RDF data, and the pre-trained embeddings can be reused in different deep neural network architectures. Later, intending to capture the most important information for each entity and make the walks less random, Cochez et al.~\cite{biased_RDF2vec} introduced biases to the walks, which led to significant improvements. Along with the edge labels, they augmented each edge with weight using twelve different strategies.  Whenever a walk arrives at a  vertex, a probability is computed for the possible edges that decide whether to take that edge. Hence, these weights bias the random walks on the graph. Although both the works by Ristoski et al. and Cochez et al.~\cite{rdf2vec,biased_RDF2vec} were evaluated for several tasks, such as entity relatedness, the approach is suitable for handling link prediction and fact prediction tasks. In OPA2Vec~\cite{opa2vec}, and Onto2Vec~\cite{onto2vec}, a similar approach as RDF2Vec was used. In Onto2Vec, Smaili et al. uses the axioms of an RDFS ontology as the corpus, that is, it treats each axiom as a sentence for training, while in OPA2Vec, Smaili et al. complements the corpus of Onto2Vec with the lexical information provided by ontology metadata, such as rdfs:comment and rdfs:label. This metadata could potentially provide valuable information about different aspects of the entities. 
The approaches were evaluated for predicting protein-protein interaction, but the embeddings could be used for graph completion as well. 
Since these methods treat each axiom as a sentence and do not utilize the interconnections offered by graph structures, it is hard to explore the entity correlations and the logical relation between axioms. Also, considering only the axioms as sentences might lead to a shortage of training corpus for small to medium-scale ontologies. 
The fundamental method that these walk-based techniques employed were further extended to other types of logic as well, for example, OWL2Vec~\cite{owl2vec}. The ontology embedding methods based on these walk-based embeddings convert the ontology to a graphical form that can capture the logical structure and inter-entity relationships. Modeling RDF(S) ontologies as graphs is easy. However, OWL ontologies include not only graph structure but also logical constructors, and entities are often augmented with richer
lexical information specified using \texttt{rdfs:label}, \texttt{rdfs:comment}, and many other built-in annotation properties. So modeling OWL axioms as graphs is a complex problem. The general assumption for such techniques is that simpler logic can easily be represented as graphs than more expressive logic, and hence OWL ontologies were projected to RDF graphs~\cite{ontology_projection} to perform walks. To create embeddings, a series of graph walks were conducted, and the sequences were collected in a document. Similar to RDF2Vec, word2vec is used to train the embeddings. But, this approach did not fully capture the logical and lexical information in the graph. Later, Chen et al. introduced OWL2Vec*~\cite{owl2vecstar}, an extension of OWL2Vec. OWL2Vec* intending to capture the structure, logical constructors involved in each axiom, and the relationship between entities across axioms, generates three documents by walking over the graph. Each document consists of a corpus that captures different aspects of the ontology (i) the graph structure (obtained from the projected RDF graph) and the logic constructors (obtained from the ontology axioms), (ii) the lexical information (obtained from the annotation axioms, comments, and definitions), and (iii) a combination of corpus from steps (i) and (ii). After the documents are created, they are combined into a single document as one corpus. Finally, OWL2Vec* uses a word embedding model to create embeddings of entities from the generated corpus. The performance was evaluated for concept subsumption and membership tasks. 

Later on, the advent of deep learning opened doors for new research directions. Hohenecker and Lukasiewicz~\cite{DL_for_OR} proposed Relational Tensor Networks (RTN), an adaptation of Recursive Neural Tensor Networks (RNTN)~\cite{RNTN} for relational learning. RNTN was initially designed  to support learning from tree-structured data. RTN is an RNTN that makes use of a bilinear tensor layer. The term \textit{relational} in an RTN emphasizes the focus on relational datasets. The intuition is that the most
critical information about individuals is hidden in their relations.
The authors start by building a Directed Acyclic Graph (DAG) representation of the input RDF ontology where directed edges represent the binary relations, and each of the vertices represents the individuals. Each individual is represented as an incidence vector that indicates the concepts they belong to. The embeddings of
the individuals are computed using the RTN model that considers
the type or the relation each individual has.  They consider two tasks, type prediction (membership of individuals to concepts) and relation prediction. The input for relation prediction is the embeddings of two individuals for which the relations are being classified. The results reported on large datasets for both
predictive performance (on concept and relationship prediction) and time consumption (for import and materialization) showed performance improvements compared to RDFox. However, to generate DAG, traditional reasoners (Apache Jena\footnote{\url{https://jena.apache.org}}, Pellet\footnote{\url{https://github.com/Complexible/pellet}}) were used. Due to this, RTN is not noise-tolerant. For learning, the
datasets (including the inferences) were divided into training, test, and validation
sets, meaning the performance is reported on the same ontology, and learning is not transferable to other domains. In order to ensure that there is sufficient data for training, only those
relations were considered that appear in at least 5\% of the total individuals. 

\begin{figure}
\centering 
     \includegraphics[width=0.80\textwidth]{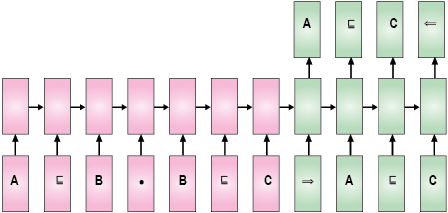}

       \caption{Seq2seq translator like setting of logical reasoning}
         \label{fig:1}
      
 \end{figure}

Recently, in an effort to emulate the reasoning capabilities of traditional logical reasoners, deep learning techniques have been utilized to learn the working of entailment rule-based RDF(S) reasoners  (see Table \ref{tab:rdfsrules} for the rules). The first and foremost requirement to work with deep neural networks (DNNs) is  that the ontology needs to be converted to a form such that it can focus more on the logical structure and learn the general working of the traditional reasoning algorithms instead of just focusing on the similarity based on the meaning or the textual representation. As shown in Figure~\ref{fig:1}, given a set of axioms (\texttt{A} $\sqsubseteq$ \texttt{B} and \texttt{B} $\sqsubseteq$ \texttt{C}), the model should be able to capture the RDFS rule involved (RDFS11 in this case) in deriving the inference (\texttt{A} $\sqsubseteq$ \texttt{C}). In a deep learning context, this setting is similar to  sequence-2-sequence machine translation models, also called the encoder-decoder model (similar to the sequential process used in many deductive reasoning algorithms), that translate text or speech from one language to another. In doing so, it needs to predict the correct substitution of words in one language for words in another. The sentences in the input and target language could be of different lengths and have completely different structures. The difference is that for language translation, the meaning of the words needs to be captured. However, logical reasoning algorithms only learn an abstraction. For example, if we replace \texttt{A} with \texttt{C$_1$}, \texttt{C} with \texttt{C$_2$}, and \texttt{D} with \texttt{C$_3$}, the reasoner only needs to determine that the RDFS11 rule pattern should be used here 
in order to infer \texttt{C$_2$} $\sqsubseteq$ \texttt{C$_3$}. So, the next series of works mostly, though they differ in technical details, apply seq-2-seq based models to capture the functioning of the logical reasoners and generalize them to analogous situations or tasks. 


With a two-fold goal of learning RDFS rules using deep learning and demonstrating their noise-tolerance capabilities, Makni and Hendler~\cite{noise-tolerant} proposed a layered graph model for RDF where the RDF graphs were represented as a sequence of graph words, and the translation-based techniques were used for graph-to-graph learning. The proposed layered graph model is based on simple directed RDF graphs where each layer has its own set of directed edges. The first phase of encoding is to create a 3D adjacency matrix (a tensor) for the input graph. Each layer in the 3D adjacency matrix corresponds to the adjacency matrix between the RDF entities linked by a single relation. These layers are then assigned an ID that represents their layout. The sequence of these IDs represents the RDF graph called “graph words”. The training set becomes a parallel corpus between the sequence of graph words of the input graph and the sequence of graph words of the inference. The algorithm then uses a seq-2-seq encoder-decoder architecture where the input RDF graph and its corresponding inference are encoded. These encoded representations are used for training. 
After training, the predicted inference graph is obtained for the embedded input test RDF graph. Finally, the inference graph is reconstructed from the translated embedding, and predictions are extracted. However, in this work, evaluation and training are done on the same RDF graph, that is, there is no learning of the general logical reasoning, and consequently, learning is not transferable to new domains. Hence, re-training will be needed for learning entity embeddings in the new ontology. They evaluated their approach on two different datasets, a synthetic dataset from LUBM and a real-world dataset that has been extracted from DBpedia. This approach can handle only noise in the ABox, and the tolerance to TBox noise is outside the scope of this work. In other words, it is based on the assumption that only assertions can potentially be noisy, and the TBox of an ontology cannot be noisy. Also, for training, they excluded every RDFS rule with only TBox axiom patterns, such as RDFS rule 5. 
For a similar setting, Makni et al.~\cite{Explainable_rdfs_reasoning} proposed a method for generating justifications for the derived conclusions. The technique was built upon~\cite{noise-tolerant} and consists of a justification model, which is a modified seq2seq model that, while training, takes two sequences (encoded RDF graph and inference graph) as the input and one sequence (justification) as the output. The inferred triples and the justification for each were generated using Apache Jena~\cite{jena}. Note that the derivations provided by Jena are summarized derivations, so the predicted justifications are also summarized. Although the approach looks promising, they do not consider large and noisy RDFS data. 
In line with~\cite{noise-tolerant}, Ebrahimi et al.~\cite{Ebrahimi2018ReasoningOR} utilized end-2-end memory networks (MemN2N)~\cite{MemN2N} to emulate deductive reasoning due to the sequential nature of the memory networks and the attention modeling that captures only relevant
information (logical axioms) necessary for the next reasoning step. The proposed approach also addresses the transferability issue by utilizing a preprocessing step consisting of normalization where they consistently rename all the entities to a predefined set of entity names. Note that the URIs which are part of the RDF/RDFS namespace are not renamed, so the learning is based only on the structural information (the type and order of constructs in each axiom) and not the actual names of the entities. The triples ($t_i$) are treated as text. The model takes a discrete set of normalized triples $t_1$, $t_2$ ... $t_n$ along with a query q and writes them all to the memory, then calculates a continuous embedding for the triples and q through multiple memory accesses. Through multiple hop attention over these continuous representations, the model classifies the query with a 'yes' or 'no', determining whether q can be inferred from the current ontology statements or not.  The results demonstrated that a consistent renaming across all the ontologies in the datasets helped perform reasoning over previously unseen RDFS knowledge graphs without retraining. However, there was a limit on the number of sampled triples. Only 1000 triples can be considered at a time. Furthermore, the model is query-based; that is, it does not compute all the inferences that may be derived from a knowledge base but provides predictions for single queries only. They frame the problem as a classification task, i.e., check whether or not the given inference is a valid inference. Later, Ebrahimi et al. explored the capabilities and limitations of neural pointer networks~\cite{PN} for the same task in a similar setting on different logics (RDF and $\mathcal{EL}^+$, discussed later in Section~\ref{el+}) and have shown significant improvements in accuracy~\cite{pointer_nw}. The model is an encoder-decoder architecture that uses attention as a pointer to choose an element of the input knowledge graph at each decoding time step. 
They claim that since pointer networks can generalize beyond the maximum lengths they were trained on, they are suitable for emulating reasoning behavior and show an improvement in performance by a huge margin. The core of the experiments comprises training a sequence-to-sequence-based model by framing the entailment problem as an input-output mapping task wherein they used two single-layer LSTMs: an LSTM encoder that converts the input sequence to a code that is fed to the generating network and the pointer LSTM that produces a vector that modulates a content-based attention mechanism over symbols in the input. The output of the attention mechanism is a softmax distribution with a dictionary size equal to the length of the input. They used two preprocessing steps: tokenization (splitting the text into meaningful chunks) and normalization (same as~\cite{Ebrahimi2018ReasoningOR}). The proposed approach is both transferable and generative, and the results demonstrated improved performance compared to the state-of-the-art. However, they only consider a small dataset for their evaluation, and the scalability aspect hasn't been explored yet.


\subsection{Description Logic $\mathcal{EL}$}
\label{el+}

The logical constructors supported by the description logic $\mathcal{EL}$ are used by several ontology developers due to their polynomial reasoning time complexity. Intending to further the neuro-symbolic reasoning research for more expressive ontologies, some recent works proposed methods to incorporate the $\mathcal{EL}$ constructors and their extensions, such as existential restriction, conjunction, and the bottom concept (see Table \ref{tab:syntax}) as embeddings. 
 
\begin{figure}[ht]
\centering 
     \includegraphics[width=0.30\textwidth]{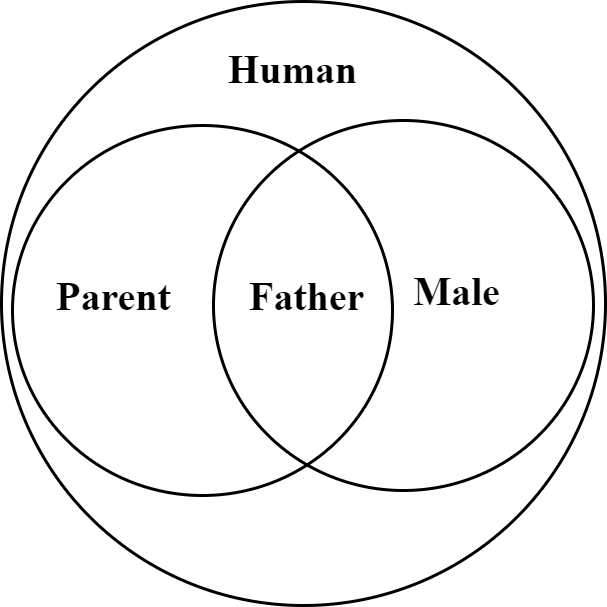}

       \caption{Entities projected as geometric shapes}
         \label{fig:4}
      
 \end{figure}

To incorporate the syntax and the underlying characteristics of the language constructs, Kulmanov et al.~\cite{ELEm} used geometric embeddings. Geometric embedding techniques have gained recent attention wherein objective (loss and score) functions for logical axioms are constructed by transforming entities in the ontology into geometric space, as shown in Figure~\ref{fig:4}. The figure shows the projection of concepts, \texttt{Parent} and \texttt{Male}, of axioms $\texttt{Parent} \sqsubseteq \texttt{Human}$, $\texttt{Male} \sqsubseteq \texttt{Human}$ and $\texttt{Parent} \sqcap \texttt{Male} \equiv \texttt{Father}$ as spheres~\cite{ELEm}. The intersection of the two spheres, \texttt{Parent} and \texttt{Male} is \texttt{Father}.

The concepts were represented as n-balls (spheres with a fixed radius) and the relations as translation vectors between the centers of each concept ball.  
TransE~\cite{Bordes_EL+}  assumes that the relationship between words can be computed by their vector difference in the embedding space. Before constructing the embeddings, the individuals were replaced with a single class axiom, and all the axioms were transformed into the standard normal form~\cite{Sweb_foundation}. For each normal form, a loss function was formulated to capture the semantics of $\mathcal{EL}^{++}$. Although the results were promising, it did not consider the RBox and certain $\mathcal{EL}^{++}$ axioms that do not translate into the embedding. Further, their use case is predicting protein-protein interactions that are modeled as a traditional link prediction task in knowledge bases. To overcome the limitations, Mondal et al.~\cite{Mondal} and Mohapatra et al.~\cite{why_settle_for_one} extended this work and showed significant improvements. Mondal et al.~\cite{Mondal} proposed EmEl++ by complementing the embeddings proposed in~\cite{ELEm} by role-inclusions and role chains (additional constructs offered by $\mathcal{EL}^{++}$). While these methods provided a new technique to model the logical structure of axioms into the geometric space, they only considered one-to-one relations that prevented them from being used in more complex DLs ontologies such as $\mathcal{SROIQ}$.  To deal with it, Mohapatra et al.~\cite{why_settle_for_one} further extended the work by Kulmanov et al. by modifying the embeddings to incorporate many-to-many relationships.  Although effective, the n-ball embeddings have two major limitations. First, the intersection of two concepts can never be a ball. For example, the intersection of two concepts (balls) \texttt{Parent} and \texttt{Male}, represents the concept \texttt{Father}, which is not a ball. Therefore, the concept
equivalence axiom \texttt{Parent} $\sqcap$ \texttt{Male} $\equiv$ \texttt{Father} cannot be captured in the embedding space. Second, simple translation does not allow for scaling the size. So, there are issues with modeling concepts of varying sizes, such as the larger concept
\texttt{Person} can not be translated into a smaller concept \texttt{Parent}. Further, the n-ball approaches regarded the axioms in ABoxes as special cases of TBox axioms and did not treat them separately. This simplification cannot fully express the logical structure of axioms. To overcome these limitations, Xiong et al. proposed BoxEL~\cite{faithful_embeddings} for embedding $\mathcal{EL}^{++}$ ontologies. They modeled the concepts in the ontology as boxes (axis-aligned hyperrectangles). The intersection of two boxes is also a box. The individuals were encoded as points inside the (concept) boxes they should belong to. The relations were modeled as the affine transformation between boxes and/or points that can capture the cases that are impossible in ball embeddings. Like ELEm, they designed several loss functions for each logical constructor in $\mathcal{EL}^{++}$ and formulated BoxEL as an optimization task. BoxEL preserves the logical structure and provides theoretical soundness guarantees in the sense that if the loss
of BoxEL embedding is 0, then the trained model is a (logical) model of the ontology. The evaluations on three ontologies for the subsumption reasoning task and predicting protein-protein interactions prove that BoxEL outperformed previous approaches. 

The other series of works by Eberhart et al.~\cite{emulation} and Ebrahimi et al.~\cite{pointer_nw}, aims to emulate the behavior of traditional (deductive) reasoners. Similar to RDF entailment rules, $\mathcal{EL}$ involves reasoning rules,  i.e., a sequence of applications of a set of pattern-matching rules that results in the addition of a new set of axioms to the ontology at each step. So the goal of these works is to learn the working of inference-rule-based reasoners, i.e., to learn the structure of reasoning patterns. Although the primary goal by Eberhart et al.~\cite{emulation} is the ontology completion wherein they predict concept inclusions and existential restrictions, this work aligns with the survey because of the shared objective of emulating logical reasoning behavior. In contrast to the other works, their approach also provides transparency of the derived conclusions. The approach works by mapping the inferences (called supports) obtained at each reasoning step in an LSTM learner. For reasoning, they use $\mathcal{EL}^{+}$  completion rules. This allows encoding of the input data in terms of ontology axioms. It also provides intermediate answers that might improve results when provided to the system. This logic data is fed into three different LSTM architectures (Deep, Piecewise, and Flat) with identical input and output dimensionalities.  The reason for choosing three architectures is to compare the behavior of the systems with and without supports. The number of LSTM cells is determined based on the maximum number of reasoning steps. The approach is able to handle noise and has shown motivating results.  Recently, Ebrahimi et al.~\cite{pointer_nw} used pointer networks (explained in Section \ref{rdf} for $\mathcal{EL}^{+}$. They claim that since pointer networks can generalize beyond the maximum lengths they were trained on, they are suitable for emulating reasoning behavior and show an improvement in performance by a huge margin. As discussed earlier, the datasets used were very small, and hence the scalability of the system has not been tested. 

\subsection{Description Logic $\mathcal{ALC}$}
The previously discussed techniques work on RDF data or slightly more expressive ontologies such as $\mathcal{EL}$, $\mathcal{EL}^{+}$, and $\mathcal{EL}^{++}$ (except~\cite{owl2vec,owl2vecstar}, which in some sense also deal with RDF graphs only wherein the OWL axioms are first projected to RDF graphs, and word2vec is applied over the sequences generated from graph walk). None of the approaches handle constructs such as negation and disjunction. These constructs, along with a few others, make the description logic $\mathcal{ALC}$ more expressive than the ones discussed so far. The neuro-symbolic reasoning work that supports $\mathcal{ALC}$ is limited but diverse in terms of the chosen methods, such as convexity~\cite{cone_semantics}, quantum logics~\cite{QuantumEmbedding}, and learning heuristics for tableau optimization~\cite{DBLP:conf/dlog/MehriH17,RL_for_tableau}.

{\"O}zg{\"u}r et al.~\cite{cone_semantics} proposed to embed concepts as convex regions in vector spaces. The approach works in two parts -- first, by constructing concepts on the axes-aligned cones (a class of convex cones) and then placing individuals on these cones. The reason behind choosing axis-aligned cones is two-fold; they are computationally feasible and are preserved under operators (intersection, polarity, and projection) that are sufficient to model constructs in the $\mathcal{ALC}$ description logics. To model unsupported constructs, set-complement, and set-union, they used a scalar product and de Morgan's laws, respectively. Since an axis-aligned cone is preserved under all these $\mathcal{ALC}$ 
 operators, the idea is that if an ontology can be projected onto an axes-aligned cone, it means that the ontology is satisfiable. The resulting embeddings can be used in concept membership prediction tasks. 
An unresolved problem in this work is whether roles can be interpreted by cones (or other feasible structures), and as a side product of defining negations by polarity, they obtain partial models. So there are individuals for which one does not know whether or not they belong to a particular concept. This is different from the approaches considered in classical embedding scenarios. Also, here, they only consider the case where the logic has been specified beforehand, not the case of investigating logic induced by the intersection and polarity operators for the arbitrary cones. 

Inspired by quantum space, Garg et al.~\cite{QuantumEmbedding} proposed Embed2Reason (E2R) that embeds an $\mathcal{ALC}$ ontology into a finite-dimensional vector space preserving the logical structure. Such an embedding, referred to as quantum embedding, satisfies the
axioms of quantum logic~\cite{quantum_mechanics} and allows one to perform logical operations (e.g., membership, negation, conjunction, disjunction, inclusion, and implication) directly over the vectors in a manner similar to the boolean logic except that distributive law does not hold true. E2R formulated an unconstrained optimization program that captures all such quantum logical (as well as regularity) constraints, and the program was solved using the Stochastic Gradient Descent (SGD) technique. The resulting embedding captured the logical input structure with good accuracy on complex deductive and predictive reasoning tasks compared to other embedding techniques. 

As discussed in Section~\ref{neuro}, the performance of the tableau-based reasoners depends a lot on the non-deterministic choices made while constructing the tableau. A series of works by Mehri et al.~\cite{DBLP:conf/dlog/MehriH17,DBLP:journals/corr/abs-1810-06617,DBLP:journals/corr/abs-1904-09443} focuses on selecting the right expansion heuristics for tableau expansion using machine learning techniques. The goal is not to emulate the tableau expansions but to learn the right expansion heuristics for intelligent decision-making at each non-deterministic rule application.
In one of these works ~\cite{DBLP:conf/dlog/MehriH17}, Mehri et al.  focused on improving semantic branching for disjunctions and its effect on backjumping, an optimization technique for backtracking. Even though semantic branching avoids redundant search space by a great deal, machine learning techniques help to decrease redundant search space exploration further by learning new orders for applying rules at each branching level. Their results show that machine learning speeds up JFact by one to two orders of magnitude. Later in~\cite{DBLP:journals/corr/abs-1810-06617}, they improved the ToDo list optimization technique
in JFact using machine learning and applying it to a complex description logic that is a syntactic variant of OWL and contains propositional logic as a very small subset. To the best of our knowledge, this is the first machine-learning approach
for improving such a rule-based optimization technique for OWL reasoners. The approach is based on an independent systematic analysis of rule orderings to uncover ontology patterns and their associated features relevant to this specific optimization technique. In continuation with this work, Mehri et al. proposed a variation~\cite{DBLP:journals/corr/abs-1904-09443} wherein they learn to choose among the built-in expansion-ordering heuristics. These learning-derived models are built based on features computed and extracted from ontologies. Therefore, the manual fine-tuning of heuristics for these reasoners can be replaced by a new ML-based approach implementing automatic fine-tuning of heuristics to make the right choice. However, the features were manually computed, and supervised learning techniques were applied to a set of ontologies to learn the best possible expansion. Also, despite a significant amount of research there is no general single sorting strategy that works best for all ontology types. That's because the ontologies vary in features, and hence the interaction and interplay of the axioms also vary. Manually designing heuristics for large and complex ontologies is nearly impossible. Therefore, it is important to develop a methodology that automatically learns to choose the most suitable expansion heuristic for each ontology. Inspired by the success of AlphaGo~\cite{alphago}, researchers started exploring reinforcement learning (RL) for intelligent human decision-making for their applications. Based on a similar idea, Singh et al.~\cite{RL_for_tableau} proposed to learn the non-deterministic expansions from scratch using deep RL such that the agent can learn the interaction and interplay between the axioms and make clever decisions whenever it has to deal with non-deterministic choices.



\subsection{OWL 2 RL}
So far, there have not been many neuro-symbolic reasoning techniques for expressive ontology profiles, such as OWL 2 RL and OWL 2 DL. Hohenecker and Lukasiewicz~\cite{patrick} developed a deep learning-based model called Recursive Reasoning Networks (RNN) for reasoning on OWL 2 RL ontologies. The approach deals with ABox and TBox separately (called factual information and ontology, respectively, in the paper). The primary goal is to train an RRN to learn the rules specified as part of TBox to answer queries related to the facts in the ABox. Since the training is relative to a particular TBox, it is independent of the ABox axioms it is provided with and can be used to answer queries over new ABox based on the same vocabulary. However, to reason over ontologies from new domains, the RRN needs to be retrained again. Hence the learning is not transferable to new domains. Since they train it on a fixed TBox, the concepts and relations are also fixed from the beginning, and that determines the number of recursive layers needed for training. 
For generating the embeddings, all the facts are first viewed as triples, where concept assertion, $C(i)$, is denoted as ($i$, memberOf, $C$), and all other relation assertions, $R(i, j)$, as $(i, R, j)$. For negative samples, such as $\neg$$C(i)$ and  $\neg$$R(i, j)$, facts are represented as ($i$, $\neg$memberOf, $C$)and ($i$, $\neg$$R$, $j$), respectively. Then, a random embedding is generated for each individual present in these triples. After this, the model updates the embeddings of the individuals by iterating over the triples. As mentioned above, the number of iterations required depends
on the concepts and relations in the respective datasets.  After these updates, each embedding is supposed to store the considered triple and encode all the inferences that are based on the triple involved. So, intuitively, a single update step conducts local reasoning based on the embeddings and the new information gained through the provided fact. Since the model goes through all the facts multiple times in several iterations, it allows for multi-hop reasoning. 
After training, predictions are made based on the obtained embeddings of all individuals. The reported results show good accuracy. However, since the updates are made in multiple steps, the method does not have the provision of generating justifications or transparency for the predicted inferences. 
In RRNs, triples are not treated as text, instead, the individuals that appear in a triple are mapped to their embeddings before providing the same to any layers of the used model. However, this means that RRNs are independent of the individual names used in a database, which makes perfect sense, as it is only the structure of a knowledge base that determines possible inferences. 
A point of difference from the other approaches~\cite{noise-tolerant,Ebrahimi2018ReasoningOR} is that the RRN is trained on a fixed ontology, i.e., the vocabulary of concepts and relations is predetermined. The model does not learn the logical structure of the axioms; hence, the learning is not
transferable to different domains. The performance is evaluated based on the query predictions. Hence the approach is not generative. 


\subsection{Summary}
\label{summary}
This section gives an overview of the neuro-symbolic reasoning techniques discussed so far from two different perspectives. In Table~\ref{tab:sot}, we categorize the existing works based on the technique. Although the technical details differ, the works can be grouped based on the high-level idea used. For each work, we give a one-line summary along with the reasoning task handled. 
In Table~\ref{tab:sot2}, inspired by Ebrahimi et al.~\cite{pointer_nw}, we highlight the key differences between the learned models in terms of the essential features that can potentially provide pointers for further research in this direction. The \textit{transfer} column indicates whether the system can adapt to ontologies in the new domain by making use of its learned model from other domains. The \textit{transparency} column indicates whether the predictions and the drawn inferences are explainable. The \textit{scale} column indicates the size of input ontologies that were used for evaluations, ranging from a few logical statements (up to 1,000 triples)~\cite{pointer_nw,Ebrahimi2018ReasoningOR}  to very large (say, up to 1 million). The \textit{performance} column indicates the accuracy of the system on the reasoning task; ``high" indicates 70\% or more, while ``low" indicates values slightly better than random guessing. Although these measures are subjective and there is no common platform where all the systems can be run together and compared on similar datasets, the goal here is just to summarize the existing neuro-symbolic reasoning techniques based on the results reported in terms of their important features. 

\begin{longtable}{|l|p{1.0cm}|p{1.7cm}|p{0.5\textwidth}|}
\hline
\textbf{Technique} &
  \textbf{Paper} &
  \textbf{Logic} &
  \textbf{Summary}  \\ \hline

\multirow{6}{*}{\begin{tabular}{c}
    \textbf{Walk-based } \\
    \textbf{Embedding}
\end{tabular}} &
  ~\cite{rdf2vec} &  RDF
   & 
   Ristoski et al. proposed RDF2Vec. Word2Vec is applied on the sequences generated using graph walks and the sub-tree RDF adaptation of the Weisfeiler-Lehman algorithm.\newline
   Task: Graph Completion (Link or Fact Prediction)
  \\ \cline{2-4} 
  & ~\cite{biased_RDF2vec} & 
   RDF & Cochez et al. extended RDF2Vec~\cite{rdf2vec} by assigning weights to the edges that bias the random graph walks and captures only the most important information. \newline
   Task: Graph Completion (Link or Fact Prediction)
   \\ \cline{2-4} 
   & ~\cite{onto2vec} & 
   RDFS & Smaili et al. proposed Onto2Vec. Each axiom is treated as a sentence and Word2Vec is applied for training. \newline
   Task: Graph Completion (Link or Fact Prediction)
   \\ \cline{2-4}  
   & ~\cite{opa2vec} & 
   RDFS &  Smaili et al. extended Onto2Vec~\cite{onto2vec} and proposed OPA2Vec by complementing the corpus with the lexical information provided by ontology metadata, e.g., rdfs:comment and rdfs:label. \newline
   Task: Graph Completion (Link or Fact Prediction)
   \\ \cline{2-4} 
 &
  ~\cite{owl2vec} &
   $\mathcal{SROIQ(D)}$ & 
    Holter et al. proposed OWL2Vec. OWL ontologies are mapped to RDF graphs and Word2Vec is applied over generated walks. \newline
   Task: Concept membership and concept subsumption
   \\ \cline{2-4}  
 &
  ~\cite{owl2vecstar} &
  $\mathcal{SROIQ(D)}$  &
   Chen et al. extended OWL2Vec~\cite{owl2vec} and proposed OWL2Vec*. The training corpus consists of three documents (structural, lexical, and combination of both) that captures the interrelation between the entities better than previous Word2Vec based methods. \newline
   Task: Concept membership and concept subsumption
   \\ \hline
\multirow{7}{*}{\begin{tabular}{c}
    \textbf{Geometric } \\
    \textbf{Embedding}
\end{tabular}}  &
 ~\cite{ELEm} &
  $\mathcal{EL}^{++}$ & 
   Kulmanov et al. proposed ELEm where the concepts were represented as n-balls and the relations as translation vectors between the centers of each concept ball.\newline Task: Subsumption
   \\\cline{2-4} 
 &
 ~\cite{Mondal} &
$\mathcal{EL}^{++}$    &
    Mondal et al. proposed  EmEL$^{++}$ by extending ELEm~\cite{ELEm} with relation inclusion and role chains. \newline Task: Subsumption
   \\ \cline{2-4} 
 &
 ~\cite{why_settle_for_one} &
 $\mathcal{EL}^{++}$    & 
   Mohapatra et al. dealt with the limitations of ELEm~\cite{ELEm} and EmEL$^{++}$~\cite{Mondal} by incorporating many-to-many relations in the embeddings.\newline Task: Subsumption
   \\ \cline{2-4} 
 &
 ~\cite{faithful_embeddings} &
 $\mathcal{EL}^{++}$     &
    Xiong et al. mapped concepts as boxes and deals with the limitations of n-ball based embeddings. \newline Task: Subsumption
      \\ \cline{2-4}  
 &
 ~\cite{cone_semantics} &
  $\mathcal{ALC}$  & 
    {\"{O}}z{\c{c}}ep et al. embed concepts as convex regions in vector spaces. \newline Task: Concept Membership
   \\ \cline{2-4}  
 &

 ~\cite{QuantumEmbedding} &
   $\mathcal{ALC}$  &
    Garg et al. proposed embeddings in the quantum space.  \newline Task: Concept Membership
   \\ \hline
\multirow{7}{*}{\begin{tabular}{c}
    \textbf{Emulating } \\
    \textbf{Logical} \\
    \textbf{Reasoning} 
\end{tabular}} &
 ~\cite{noise-tolerant} &
  RDFS & 
  Makni and Hendler used an encoder-decoder architecture for translating the embedding of the input RDF graph to the embedding of the corresponding inference graph. \newline Task: Entailment Reasoning 
  \\ \cline{2-4} 
 &
 ~\cite{Ebrahimi2018ReasoningOR} &
  RDFS & 
Ebrahimi et al. explored the capabilities of end-2-end memory networks (MemN2N). Use of normalized embeddings support transfer.  \newline Task: Query-based classification
   \\ \cline{2-4}  
 &
 ~\cite{pointer_nw} &
  RDFS and $\mathcal{EL}^{+}$  
   & Ebrahimi et al. utilized pointer networks for learning the sequential application of inference rules used in many deductive reasoning algorithms. \newline Task: Entailment Reasoning
   \\ \cline{2-4} 
 &
 ~\cite{patrick} &
   OWL 2 RL & Hohenecker and Lukasiewicz developed a deep learning-based model called Recursive Reasoning Networks (RNN).\newline Task: Entailment Reasoning
   \\ \cline{2-4} 
  &
 ~\cite{emulation} &
  $\mathcal{EL}^{+}$  & Eberhart et al. utilized LSTMs to learn the mapping of the inferences obtained at each reasoning step with the axioms in the ontology.\newline Task: Ontology completion (predict concept inclusions and existential restrictions)
   \\ \cline{2-4} 
    &
 ~\cite{Explainable_rdfs_reasoning} &
   RDFS &  
  Makni et al. built upon the work by Makni and Hendler~\cite{noise-tolerant} for generating explanations for the derived conclusions by taking the RDF graph and inferred triples as input and the explanations as the target. \newline Task: Entailment Reasoning with summarized explanations
   \\ \cline{2-4} 
 &
 ~\cite{DL_for_OR} &
   RDF & 
    Hohenecker and Lukasiewicz proposed Relational Tensor Network (RTN). Embeddings of the individuals are computed by applying RTNs on the Directed Acyclic Graph representation of the ontology (including the inferences). \newline Task: Concept Membership and Relation prediction
   \\ \hline
\multirow{2}{*}{\begin{tabular}{c}
    \textbf{Learning } \\
    \textbf{Heuristics} \\
    \textbf{for} \\
      \textbf{Tableau}
\end{tabular}} &
  \begin{tabular}[]{@{}l@{}}~\cite{DBLP:conf/dlog/MehriH17} \\~\cite{DBLP:journals/corr/abs-1810-06617}\\~\cite{DBLP:journals/corr/abs-1904-09443}\end{tabular} & $\mathcal{ALC}$  &  Mehri et al. used machine learning to select
the right heuristics for tableau expansion. The learning-derived model is built based on manually designed features that are computed and extracted from ontologies.  \newline Task: Satisfiability \\ \cline{2-4} 
 &
 ~\cite{RL_for_tableau} &
   $\mathcal{ALC}$ &
   Inspired by AlphaGo~\cite{alphago}, Singh et al. propose to use RL for learning the heuristics for non-deterministic tableau expansions from scratch. \newline Task: Consistency Checking 
   \\ \hline

 \caption{Summary of the state-of-the-art neuro-symbolic reasoning techniques over RDF and Description Logic ontologies}
\label{tab:sot}
\end{longtable}

\begin{table}[ht]
\begin{center}
\begin{tabular}{|l|l|l|l|l|l|}
\hline

  \textbf{Paper} &
  \textbf{Logic} &
  \textbf{Transfer} &
  \textbf{Transparency} &
  \textbf{Scalability} & \textbf{Performance}  \\ \hline

 ~\cite{rdf2vec} &  RDF & no & no & low & moderate

  \\ \cline{1-6}  

  ~\cite{biased_RDF2vec} & 
   RDF & no & no & low & moderate

   \\ \cline{1-6} 
   ~\cite{onto2vec} & 
   RDFS & no & no & low & moderate

   \\ \cline{1-6} 
   ~\cite{opa2vec} & 
   RDFS & no & no & low & moderate

   \\ \cline{1-6} 
 
 ~\cite{owl2vec} &
   $\mathcal{SROIQ(D)}$ &no & no & low & moderate
   \\ \cline{1-6}  
 
 ~\cite{owl2vecstar} &
  $\mathcal{SROIQ(D)}$  &no & no & low & moderate
   \\ \hline

 ~\cite{ELEm} &
  $\mathcal{EL}^{++}$ & no & no & low & moderate
   \\\cline{1-6}  
 
 ~\cite{Mondal} &
$\mathcal{EL}^{++}$   & no & no & low & moderate
   \\ \cline{1-6} 
 
 ~\cite{why_settle_for_one} &
 $\mathcal{EL}^{++}$   & no & no & low & moderate

   \\ \cline{1-6} 
 
 ~\cite{faithful_embeddings} &
 $\mathcal{EL}^{++}$     & no & no & low & high
      \\\cline{1-6}  
 
 ~\cite{cone_semantics} &
  $\mathcal{ALC}$  & no & no & low & moderate
   \\ \cline{1-6}  
 
 ~\cite{QuantumEmbedding} &
   $\mathcal{ALC}$  & no & no & low & moderate
   \\ \hline

 ~\cite{noise-tolerant} &
  RDFS & no & no & low & high \\ \cline{1-6} 
 
 ~\cite{Ebrahimi2018ReasoningOR} &
  RDFS & yes & no & moderate & high
   \\ \cline{1-6} 
 
 ~\cite{pointer_nw} &
  RDFS and $\mathcal{EL}^{+}$   & yes & no & no & high
   \\ \cline{1-6}  
 
 ~\cite{patrick} &
   OWL 2 RL & no &  no & low & high
   \\\cline{1-6} 
  
 ~\cite{emulation} &
  $\mathcal{EL}^{+}$ & yes & yes & moderate & low
   \\ \cline{1-6}  
    
 ~\cite{Explainable_rdfs_reasoning} &
   RDFS & no &  yes & low & high
   
   \\ \cline{1-6} 
 
 ~\cite{DL_for_OR} &
   RDF & no & no & high & high
   \\ \hline

  \begin{tabular}[c]{@{}l@{}}~\cite{DBLP:conf/dlog/MehriH17} \\~\cite{DBLP:journals/corr/abs-1810-06617}\\~\cite{DBLP:journals/corr/abs-1904-09443}\end{tabular} &
 $\mathcal{ALC}$  & yes & yes & moderate & high
  \textbf{} \\\cline{1-6}  
 
 ~\cite{RL_for_tableau} &
   $\mathcal{ALC}$ & yes & yes & yes & na
   \\ \hline
\end{tabular}
\end{center}
 \caption{Overview of the neuro-symbolic reasoning techniques from the point of view of key evaluation criteria: Transfer (yes or no), Transparency (yes or no), Scalability(low, moderate and high), and Performance(low, moderate and high). na implies not applicable.}
         \label{tab:sot2}
\end{table}

\section{Other Related Efforts}
\label{other}
Although there are several important developments in the neuro-symbolic reasoning space (discussed in the earlier sections) and advancements continue to happen, there still is a lack of availability of
\begin{enumerate}
    \item \textbf{Test Cases.} Ontologies that vary in terms of several parameters, such as ontology size and axiom types.
    \item \textbf{Common Infrastructure and Experiment Design.} A platform where all the existing systems can be run together and compared against some performance metrics. 
\end{enumerate} 
One way to bridge this gap and foster a strong research community is to hold challenges. This could further help the new ontology and system developers to check the hardness of the ontologies and find the performance bottlenecks of their systems. With this aim and inspired by ORE~\cite{ORE}, two editions of the Semantic Reasoning Evaluation Challenge (SemREC)\footnote{\url{https://semrec.github.io/}} were held so far. In the first edition of SemREC\footnote{\url{https://semrec.github.io/SemREC2021.html}}~\cite{semrec_proceedings}, co-located with the 20th International Semantic Web Conference (ISWC 2021\footnote{\url{https://iswc2021.semanticweb.org/}}), submissions were invited across three tracks. In the first track, ontologies that challenge the reasoners can be submitted. The second and third track is for RDFS and description logic reasoners that make use of traditional and neuro-symbolic techniques, respectively. The second edition of SemREC was co-located with the 21st International Semantic Web Conference (ISWC 2022\footnote{\url{https://iswc2022.semanticweb.org/}}). Since the research in conventional reasoning optimization has stagnated, in the second edition of SemREC, the second track from the first edition was dropped. The track on neuro-symbolic reasoning systems has been expanded to include multi-hop reasoners and inductive reasoning techniques. Across the two editions, there were three submissions related to neuro-symbolic reasoning~\cite{biswesh_semrec,sulogna_semrec,shervin_semrec}. The system submitted by Mohapatra et al.~\cite{biswesh_semrec} is the same as~\cite{why_settle_for_one} that was discussed in section~\ref{sota}. They
provided an effective way to capture the many-to-many relations between the concepts and built on top of~\cite{ELEm,Mondal}. Chowdhury et al.~\cite{sulogna_semrec} used an end-to-end Memory Network (MemN2N)~\cite{MemN2N} with attention that captures the most relevant information to conduct logical reasoning. Mehryar et al.~\cite{shervin_semrec} submitted aTransE, an extension of TransE and rTransE to make subsumption and instance checking reasoning possible, whereby reasoning can take place in the vector space by leveraging transitive relations. They further improved the quality of embeddings using reinforcement learning, where they generated multi-hop samples using a policy network. The
agent is a neural network that takes as input an entity vector embedding (i.e.,
state) and outputs a relationship (i.e., action). The use of RL helped find more
meaningful and longer sequences of translations. All the submissions were evaluated on ontologies from different sources such as the ORE benchmark framework~\cite{ORE}, OWL2Bench~\cite{owl2bench}, and CaLiGraph~\cite{caligraph}. The ORE benchmark framework\footnote{\url{https://github.com/ykazakov/ore-2015-competition-framework}} is an open-source java-based framework that was a part of the OWL Reasoner Evaluation (ORE) Competition. The competition evaluated the performance of OWL 2 complaint reasoners over several different OWL 2 EL and OWL 2 DL ontologies. But, the performance evaluation in the context of varying sizes of an ontology was not considered. The ORE competition corpus can be used with the framework for reasoner evaluation. The framework evaluates the reasoners on three reasoning tasks -- consistency checking, classification, and realization. The framework does not include the evaluation of the SPARQL query engines (with OWL reasoning support) with respect to the coverage of OWL 2 constructs and scalability.  OWL2Bench, on the other hand, tests the OWL 2 reasoners in terms of their coverage, scalability, and query performance. It is an extension of the well-known University Ontology Benchmark (UOBM)~\cite{UOBM}. OWL2Bench consists of the following -- TBox axioms for each of the four OWL 2 profiles (EL, QL, RL, and DL), a synthetic ABox axiom generator that can generate axioms of arbitrary size, and a set of SPARQL queries that involve reasoning over the OWL 2 language constructs. CaLiGraph ontologies were the winners of task 1 (challenging ontology submission) of SemREC 2021. CaLiGraph is a large-scale cross-domain knowledge graph generated from Wikipedia by exploiting the
category system, list pages, and other list structures in Wikipedia, containing more than 15 million typed
entities and around 10 million relation assertions. Other than the knowledge graphs such as DBpedia~\cite{dbpedia} and YAGO~\cite{yago}, whose ontologies are comparably simplistic, CaLiGraph has a rich ontology comprising more than 200,000 class restrictions. Those two properties, a large ABox and a rich ontology, make it an interesting challenge for benchmarking reasoners. Since all the submissions to the neuro-symbolic reasoner track in the second edition of SemREC belonged to a different category with respect to the task on which the system was evaluated on, it was not possible to compare them. For instance, Sulogna et al.~\cite{sulogna_semrec} submitted a system that supports the \textit{transfer} property. The evaluation was performed on a query-based task, where they obtained good results on the CaliGraph dataset without retraining. But their system supports only RDFS axioms, and there was a limit on the number of axioms (1000 triples) used for training. For that, they randomly divided each ontology into multiple datasets of 1000 triples each. The other submission by Shervin et al.~\cite{shervin_semrec} was designed for the link prediction task (subclass and type of relations), and the metric used was Hits@n. 


The growth of neuro-symbolic and conventional ontology reasoners largely depends on the datasets that are available for evaluating them. Most reasoner evaluations are performed on a subset of several thousands of ontologies that are available across repositories such as the DBpedia~\cite{dbpedia}, YAGO~\cite{yago}, Wikidata~\cite{wikidata}, Claros\footnote{\url{https://www.clarosnet.org}}, NCBO Bioportal\footnote{\url{https://bioportal.bioontology.org/}}, and AgroPortal\footnote{\url{http://agroportal.lirmm.fr/}}. However, such an approach is inflexible as most ontologies involve only
a limited set of OWL constructs and arbitrarily large and complex ontologies are seldom available that can be used to test the limits of systems being benchmarked. Similarly, existing synthetic benchmarks, such as LUBM~\cite{LUBM}, UOBM~\cite{UOBM}, OntoBench~\cite{Ontobench}, ORE, and OWL2Bench, also do not have the flexibility to generate ontologies that vary in terms of \textit{TBox} size and language constructs. To evaluate the reasoners, it is important to have a benchmark that can generate ontologies depending on the task at hand. For example, if the performance of a neuro-symbolic reasoner needs to be evaluated or compared with another reasoner on the concept subsumption task for $\mathcal{EL}^{++}$ ontologies, we need to have a benchmark that can generate a large number of ontologies that vary in size with respect to different OWL constructs.~\cite{owl2bench2}, an extension of OWL2Bench describes an ongoing effort towards building such a customizable ontology benchmark for OWL 2 reasoners.

\section{Conclusion}
\label{conclusion}
In this chapter, we gave an overview of the neuro-symbolic reasoning techniques over RDF and description logic ontologies. We now answer the research questions posed in Section \ref{sota}. 
Although the body of published work (Table~\ref{tab:sot} and Table~\ref{tab:sot2}) is relatively small, it is obvious from the results that deductive reasoning is a very hard task for deep learning with increasing hardness, especially for more complex logic and scalability issues. It can be observed that, under certain provisions, even the best reasoning models based on machine learning are still not in a position to compete with their symbolic counterparts.  To an extent, it has been shown that the developed neuro-symbolic techniques are noise-resistant, but they need negative samples to be added to the datasets, which is mostly done manually or randomly. 
Despite all the efforts, existing neuro-symbolic reasoners are still unable to do the following.  
 \begin{itemize}
\item{Perform full reasoning over expressive ontologies such as OWL 2 DL. The existing methods have limitations in learning complex ontology axioms. Most techniques used simpler and less expressive ontology profiles. However, interestingly, the body of work indicates the potential and feasibility of deep learning techniques for reasoning over more expressive ontologies. }
\item{Perform reasoning over new ontologies (domain-independent reasoning) with high precision and recall. Although some techniques support the \textit{transfer} property, the method used was based on the consistent renaming of all the entities in the ontologies (each entity is assigned a unique numeric identifier). But this does not solve the out-of-vocabulary issue because the highest number assigned to an entity depends on the size of the largest ontology used for training. The size of the test ontology could never be larger than that. However, the initial results are interesting and highly motivating for further work in that direction.}
\end{itemize}

A major aspect that the researchers in the domain have probably overlooked is the unavailability of the datasets and a common evaluation framework. Clearly, all the techniques discussed indicate the scope for research and improvements in this direction. It would have been great if all the published or yet-to-be-published works could have been trained and tested on the same datasets using the same resources. From the results reported in the summary table, if the scalability and performance are high, without any transfer, then the approach is clearly unsuitable for the deductive reasoning goal. On the other hand, some techniques support the transfer property but fail to consider the dataset variety (varying language constructs and dataset sizes). In order to make progress, firstly, it is very important to incorporate the datasets in evaluations that vary in terms of several parameters, and these datasets should become the standard for the neuro-symbolic reasoning systems. Secondly, automatic learning should be incorporated. Recently, reinforcement learning for FOL~\cite{fol_for_rl} has shown tremendous improvements where the systems learn to perform reasoning from scratch. For instance, in the case of reasoners that use inference-rule based methods, instead of providing the inferences (output) beforehand and then mapping the input and the output sequences using seq-2-seq translators, an RL agent can be provided with all the ontology axioms, along with the inference rules. The RL agent can then start randomly applying the inference rules on the given set of axioms. The rewards can be designed so that the agent learns to reach the desired conclusion in the least number of steps. Learning from scratch has an added advantage over labeled/supervised methods, where neural networks learn by exploring several possible moves that could lead to either right or wrong decisions (human-like learning). Something similar can also be explored for the tableau based reasoners. Extending the architectures discussed here with learning from scratch would enable transparency and explanation generation alongside the computed predictions. 


\bibliographystyle{tfnlm}
\bibliography{main}
\end{document}